\documentclass[10pt]{article} 
\usepackage{colacl}

\input{psfig}

\title{Anchoring a Lexicalized Tree-Adjoining Grammar for Discourse}
\author{Bonnie Lynn Webber \and Aravind K. Joshi\\
Department of Computer \& Information Science\\
University of Pennsylvania\\
Philadelphia PA USA 19104-6389\\
$[$bonnie, joshi$]$@central.cis.upenn.edu}

\begin{document}

\maketitle

\begin{abstract}
We here explore a ``fully'' lexicalized Tree-Adjoining Grammar for discourse
that takes the basic elements of a (monologic) discourse to be not simply
clauses, but larger {\em structures} that are {\em anchored} on
variously realized discourse cues. This link with intra-sentential grammar
suggests an account for different patterns of discourse cues, while the
different structures and operations suggest three separate sources
for elements of discourse meaning: (1) a compositional semantics
tied to the basic trees and operations; (2) a presuppositional semantics
carried by cue phrases that freely adjoin to trees; and (3) general
inference, that draws additional, defeasible conclusions that flesh
out what is conveyed compositionally.
\end{abstract}

\section{Introduction}
In the past few years, researchers interested in accounting for how
elements combine in a discourse, have taken to using the adjoining
operation found in Tree-Adjoining Grammar (TAG)
\cite{gardent94,gardent97,pol-van96,schilder97a,vandenberg96,webb91}.
More recently, Cristea and Webber \shortcite{cw97} have argued that a
Tree-Adjoining Grammar for discourse would also need the
{\em substitution} operation found in a lexicalized TAG
\cite{schabes90}. Here we move
further and explore a {\em fully} lexicalized TAG for discourse, allowing
us to examine how the insights of lexicalized grammars -- that the basic
elements of a clause are not simply words, but {\em structures} that reflect
a word's role and syntactic/semantic scope -- carry over to discourse.
We show how this suggests explanations for such phenomena as the
following:
\begin{itemize}
\addtolength{\itemsep}{-5pt}
\item that arguments of a coherence relation can be stretched
``long distance'' by intervening material;
\item that multiple discourse cues can appear in a single sentence or
even a single clause;
\item that when discourse cues appear in the middle of clauses, they
contribute to coherence in more specific ways;
\item that coherence relations can vary in how and when they are
realized lexically.
\end{itemize}

One way of understanding the current work is that it extrapolates from
lexically-based views of how structure and meaning are associated
within a sentence to how aspects of discourse structure
and meaning might be associated in similar ways. While the idea that
discourse-level mechanisms might
resemble intra-sentential mechanisms has long been an undercurrent
within discourse research, we have come to believe that the framework
of lexicalized grammar can be effectively used to demonstrate
the validity of this intuition. While we present the ideas in terms
of one well-known formalism -- Lexicalized
TAG -- other lexicalized formalisms such as CCG \cite{steedman96a}
might prove equally useful for expressing the same theoretical
insights and implementing them for discourse generation and/or
interpretation.

A superficial reading of the current proposal might
suggest that it is merely a simple embedding
of RST \cite{mt88} in TAG. That would be incorrect.
First, the primary feature of a fully lexicalized TAG is that each
elementary tree in the grammar has an {\em anchor} that indexes the
tree and defines its syntactic/semantic scope.
Here, we posit a set of {\em initial} (non-recursive)
trees, whose anchor is a discourse cue. Structurally, some
initial trees resemble the nucleus-satellite structures of
RST, and some, its joint schema. But the resemblence is only
superficial, as initial trees have a purely compositional semantics
that makes no assumptions about what the speaker is trying to
do.\footnote{The LTAG formalism itself allows an elementary tree
to be associated with a meaning that is not compositional 
with respect to its sub-parts. This is used, for example, for
associating meaning with syntactically-flexible idioms. However,
we have not found the need to exploit this possibility for
discourse, though we leave open the possibility.}

Secondly, there is a single {\em auxiliary} tree
whose semantics corresponds simply to continuing the
description conveyed by the structure
to which it is adjoined. Any additional inferences that a
listener draws from the resulting
adjacency are defeasible, and may be cancelled
or corrected by material in the subsequent discourse.
Our proposal thus factors the combinability of elementary discourse
clauses from inferences that may then be drawn, thus providing a tool
for sorting out different semantic processes in discourse, instead of
lumping them into a single category. 
Many of these inferences have been given the status of
{\em discourse relations} in RST. However, we argue in
Section~\ref{compinf:sec} that one can gain from
distinguishing what is derived compositionally
from what is derived inferentially.

Thirdly, there are auxiliary trees for other discourse cues,
that can adjoin to either initial trees or auxiliary trees.
These discourse cues contribute meaning (and coherence) through
their presuppositions or assertions or both.
They can thereby serve to constrain the range of
inferences that a listener might draw when a description is
extended, limiting them to ones compatible with the contribution of
the discourse cue. Similarly, a discourse cue adjoined to an
initial tree can either further specify the compositional meaning
of the related units or constrain how that initial tree can be used in
extending another description. This will explain how several discourse
cues can appear in the same sentence or even the same clause, each
contributing to either the compositional or presuppositional semantics
of the discourse (Section~\ref{aux:sec}).

This is still a ``work in progress'', with many open questions.
However, it may still pique the interest of two historically distinct
groups: it may stimulate people working on syntax to look beyond the
clause for phenomena familiar to them within it, while it may
help people working on discourse to ground their claims and insights
in more traditional varieties of linguistic formalisms.

\section{Elements of a Lexicalized TAG for Discourse}

A lexicalized TAG begins with the notion of a lexical {\em anchor},
which can have one or more associated tree structures. For example,
the verb {\em likes} anchors one tree corresponding to
{\em John likes apples}, another corresponding to the topicalized
construction {\em Apples John likes}, and a third corresponding to
the passive construction {\em Apples are liked by John}. All in all,
there is a tree for each minimal syntactic construction in which
{\em likes} can appear, all sharing the same predicate-argument structure.
This syntactic/semantic encapsulation is possible because of the extended
domain of locality of LTAG. Trees in such a {\em tree family} may differ
in being licensed by different states of the discourse (i.e.,
{\em information structure} \cite{steedman96b}).

A lexicalized TAG contains two kinds of elementary trees: {\em initial}
(non-recursive) trees that reflect basic functor-argument dependencies
and {\em auxiliary} trees that introduce recursion
and allow elementary trees to be modified and/or elaborated. In
our lexicalized discourse TAG, we have so far found the need to posit
only two types of initial tree families (Section~\ref{elem:sec})
and two types of auxiliary trees (Section~\ref{aux:sec}).
While the resulting grammar is thus very simple -- only one type, only
binary predicates -- it so far appears expressively adequate.

\subsection{Initial Trees}
\label{elem:sec}

\begin{figure*}
\centerline{\psfig{figure=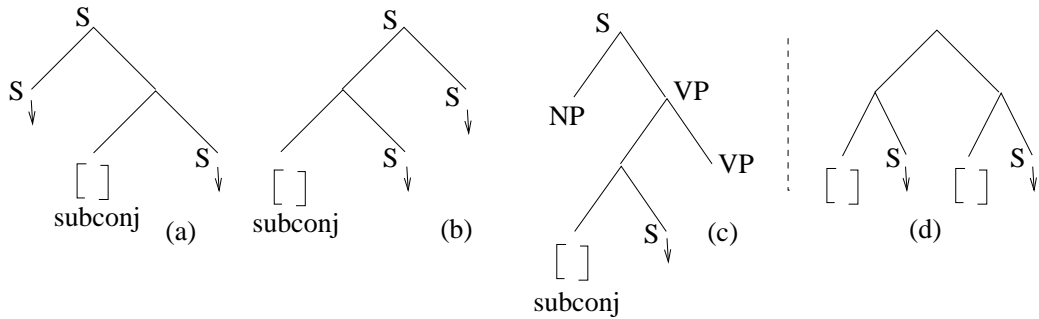}}
\caption{Initial Trees (a-c) belong to the tree family
for a subordinate conjunction. The
symbol $\downarrow$ indicates a substitution site, while $[~]$ stands for
a particular subordinate conjunction and its feature structure.
(d) is the initial tree for a parallel construction.}
\label{subord:fig}
\end{figure*}

Subordinate conjunctions are one major class of discourse
cues, and clause-level LTAG already provides an account of
subordinate clauses with overt subordinate conjunctions. 
Its  ``verb-centric'' account in \cite{tech-rept95} appropriately
treats subordinate clauses as {\em adjuncts} -- i.e., auxiliary
trees. However, from a discourse perspective, we treat
two clauses connected by a subordinate conjunction as an
{\em initial tree} whose compositional semantics reflect the
subordinate conjunction as predicate (or ``functor'') and the
clauses as arguments. There is an initial tree for each minimal
structural pattern of main clause and subordinate clause, including
those shown in Figure~\ref{subord:fig}. All such trees share the
same predicate (functor) argument structure. As in clause-level
tree families, each pattern may have different preconditions on its
use that reflect the current state of the discourse (i.e.,
{\em information structure}).
For example, it has been noted that a ``when'' clause in
initial position presupposes that the situation described
therein is in the hearer's {\em discourse model} (or can be
so accommodated), while a ``when'' clause coming after the
main clause is not so constrained.

In Section~\ref{feat:sec}, we discuss reasons for taking the lexical
anchors of these initial trees to be feature structures that may
correspond to one or more subordinate conjunctions such as ``if''
and ``when''. Here we just take them to be specific lexical items.

Now one reason for taking something to be an initial tree
is that it has local dependencies that can be stretched
long-distance. For example, the dependency
between {\em apples} and {\em likes} in both {\em John likes apples} and
{\em apples John likes} is localized in all the trees for {\em likes}.
It can be stretched, however, long-distance as in
{\em Apples, Bill thinks John may like}. In \cite{cw97}, we have
shown that the same long-distance stretching of dependencies
occurs with both subordinate clauses (Ex.~1) and parallel
constructions (Ex.~2) -- e.g.
\begin{enumerate}
\item a. Although John is very generous,\\
~b. giving money to whoever asks,\\
~c. when you actually need it,\\
~d. you'll see that he's a bugger to find.

\item a. On the one hand, John is very generous.\\
~b.~For example, suppose you needed some money.\\
~c.~You would just have to ask for it.\\
~d.~On the other hand, he's a bugger to find.
\end{enumerate}
Thus here we also posit an initial tree for parallel structures
(Figure~\ref{subord:fig}d). Since there are different ways in
which entities are taken to be parallel, we currently
assume a different initial tree for {\em contrast}
(``on the one hand'' \ldots ``on the other hand'' \ldots),
{\em disjunction} (``either'' \ldots ``or'' \ldots),
{\em addition} (``not only'' \ldots ``but also'' \ldots),
and {\em concession} (``admittedly'' \ldots ``but''
\ldots). Such trees have a {\em pair} of anchors with two main
properties.

The first is that their lexical realization seems optional.
In contrastive cases, a medial anchor such as
``on the other hand'' often appears lexicalized without an
initial phrase such as ``on the one hand''.
In fact, there are more cases of this in the Brown Corpus than of
the two appearing together.
Also optional is the realization of the initial
anchor in disjunction (omitting ``either''), addition (omitting
``not only''), and concession (omitting ``admittedly'').
But we have recently noted cases where only the initial anchor is
realized lexically but not the medial anchor, although this is less
common:
\begin{itemize}
\item[] Not only have they $[$Rauschenberg's blueprints$]$ survived. The
process of their creation was recorded by {\em Life} magazine in
April 1951. ({\em New York Review of Books}, 6 November 1997, p.8)
\end{itemize}

The second property is that the medial anchor appears realizable
in multiple ways. Cristea and Webber
\shortcite{cw97} report that, of
the eleven instances of ``on the one hand'' found in the Brown Corpus,
four have their contrasting item cued by something other than
``on the other (hand)'' -- including ``but'' and ``at the same time'':
\begin{enumerate}
\addtolength{\itemsep}{-5pt}
\setcounter{enumi}{2}
\item On the one hand, the Public
Health Service declared as recently as October 26 that present
radiation levels resulting from the Soviet shots ``do not warrant
undue public concern'' or any action to limit the intake of
radioactive substances by individuals or large population groups
anywhere in the Aj.  {\bf But} the PHS conceded \ldots.({\em cb21})

\item Brooklyn College students have an ambivalent attitude toward
their school.  On the one hand, there is a sense of not
having moved beyond the ambiance of their high school.  This is
particularly acute for those who attended Midwood High School directly
across the street from Brooklyn College.  They have a sense of
marginality at being denied that special badge of status, the
out-of-town school.  {\bf At the same time}, there is a good deal of
self-congratulation at attending a good college \ldots ({\em cf25})
\end{enumerate}
Other examples occur with ``on the other extreme''
and ``at the other extreme'' -- cf.
\begin{enumerate}
\setcounter{enumi}{4}
\item On the one hand we have the ``All you have to do is buy it'' brigade
who seem to think the only problem is that we haven't gone and ``done it''.
{\bf On the other extreme} there are groups who think if it has been
explored theoretically then it's been done.  
\end{enumerate}
In Section~\ref{feat:sec}, we will argue that both these properties
can be accommodated by treating the lexical anchors
of these initial trees as feature structures.

\subsection{Auxiliary Trees}
\label{aux:sec}

\begin{figure*}
\centerline{\psfig{figure=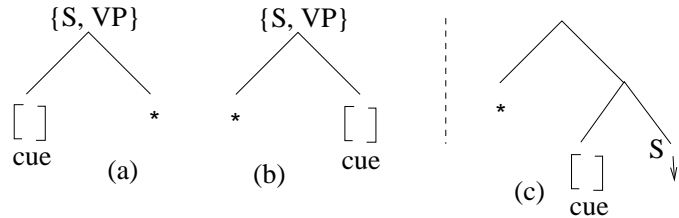}}
\caption{{\bf Auxiliary Trees}. (a) and (b) are auxiliary trees in the
tree family for adverbial discourse cues, which serve to modify or
constrain the relation holding between discourse units. Trees in
the family may be rooted on S or VP. The symbol $*$ indicates the
{\em foot node} of the auxiliary tree. (c) is the auxiliary tree for
basic elaboration.}
\label{modtree:fig}
\end{figure*}

Discourse cues other than subordinate conjunctions are either adverbs
(adverbial phrases) or conjunctions. In XTAG \shortcite{tech-rept95},
adverbials are handled as simple auxiliary trees
(Figure~\ref{modtree:fig}a-b).
We do the same here, associating each cue with a feature structure
that indicates its semantic properties for reasons to be discussed
in Section~\ref{feat:sec}.  Semantically,
such auxiliary trees can be used to elaborate or clarify the
discourse relation holding between two discourse units. This may
result in the phenomenon of there being
more than one discourse cue in a sentence, as in
\begin{enumerate}
\addtolength{\itemsep}{-5pt}
\setcounter{enumi}{5}
\item Stephen Brook, in {\em Class: Knowing your place in modern
Britain}, begins promisingly with the proposition that ``class
distinction and class consciousness -- they are both with us''.\ldots
Brook {\bf then}, {\bf however}, runs into trouble because
he feels obliged to provide a theory \ldots.

\item {\bf Although} the episodic construction of the book often makes it
difficult to follow, it {\bf nevertheless} makes devastating reading.
\end{enumerate}
We will discuss the semantics of such examples shortly, and also the
conjunctions ``so'' and ``but''.

As noted in Section~1, an auxiliary tree (here shown in
Figure~\ref{modtree:fig}c) is used to adjoin to a
structure and continue the description of the entity
(object, event, situation, state, etc.) that it
conveys. Such a tree would be used in the derivation
of a simple discourse such as:
\begin{enumerate}
\setcounter{enumi}{7}
\addtolength{\itemsep}{-5pt}
\item John went to the zoo. He took his cell phone with him.
\end{enumerate}
Here, the foot node of the auxiliary tree in Figure~\ref{modtree:fig}c
would be adjoined to the root of the tree for the first clause,
and its substitution site filled by the tree for the second.
The tree's anchor may have no lexical realization (as here, between main
clauses), or it may be realized by ``and'' (as in embedded clauses --
e.g. ``Fred believes that John went to the zoo and that he took
his cell phone with him''). The compositional meaning associated with
adjoining this tree is simply that the meaning of the second
clause continues the description of the same entity as the first.
Other aspects of meaning -- such as there being a causal connection
or temporal relation between its sub-parts, or an evidential
relation between them -- would be derived inferentially,
and hence possibly be found inconsistent, given the subsequent
discourse.

When an adverbial discourse cue is adjoined to a clause, it can
constrain how the clause can be interpreted as continuing
the already-started description -- for example,
\begin{enumerate}
\setcounter{enumi}{8}
\addtolength{\itemsep}{-5pt}
\item John went to the zoo. {\bf However}, he took his cell phone with
him.
\end{enumerate}
Following Knott and others \shortcite{knott96},
we take the semantics of such discourse cues to be
{\em presuppositional}. For example, according to Knott,
``however'' presupposes the existence
of a (shared) defeasible rule, some or all of whose antecedents
are licensed by the previous discourse, but which fails to hold
either because the clause so marked contradicts either the conclusion
or an antecedent. In Example~(9), the defeasible rule might
be something like
\begin{quote}
When people go to the zoo, they leave their work behind.
\end{quote}
So the clause marked by ``however'' in (9) both continues the
description of the event of John's going to the zoo (compositional
semantics) and conveys that the above rule fails to hold because
its conclusion is contradicted (presuppositional semantics).

Of course, since these relation-modifying auxiliary trees are
adverbials, they can, at least in English, be adjoined elsewhere in
the structure, not just at the anchor -- e.g.
\begin{enumerate}
\setcounter{enumi}{9}
\addtolength{\itemsep}{-5pt}
\item Cracked and broken plastic tableware will attract germs, so it should
be thrown away, never mended. Plastic furniture and toys, {\bf however}, can
be repaired successfully with the appropriate adhesive.
\end{enumerate}
We speculate that such medially-occuring discourse cues (of which we
are acquiring a growing corpus of naturally-occuring examples)
occur at the boundary between a sentence's topic or {\em theme}
(i.e., the question under discussion) and its comment or {\em rheme}
(i.e., the contribution made towards that question)
\cite{steedman96b}. There are then three possibilities:
\begin{itemize}
\addtolength{\itemsep}{-5pt}
\item The cue merely makes the boundary explicit, while its
presuppositional semantics remains tied to the proposition as a
whole;
\item The presuppositional semantics of the cue is grounded in
whichever informational unit (theme or rheme) occurs to its left;
\item The presuppositional semantics of the cue is grounded in
the theme (wherever it occurs), specifying how the theme
links to the discourse (i.e., {\em how} it is the question under
discussion).
\end{itemize}
Deciding among these alternatives requires more time for thought
and analysis of both constructed and such ``naturally-occuring'' examples
as
\begin{enumerate}
\addtolength{\itemsep}{-5pt}
\setcounter{enumi}{10}
\item A soldering iron is a much more specialized tool, which you will
rarely need. {\bf If} the occasion does arise when you need to solder two
pieces of metal together, {\bf however}, choose a large electric soldering
iron with a tapered head.
\end{enumerate}
and Examples~(6) and (7) above. In (11), the subordinate clause
itself is the theme. Such examples as (7) and (11) call into
question RST's assumption that {\em satellites}, which these subordinate
clauses would be taken to be, can be omitted without a great change in
meaning to a discourse. These certainly cannot.

Another open question (but more of a technical detail) is the
appropriate handling of conjunctions such as ``so'' and ``but''. On
the one hand, their semantics can best be seen as presuppositional --
presupposing a defeasible rule grounded in the previous discourse
that succeeds in the case of ``so'' and fails in the case of ``but''
\cite{knott96}. On the other hand, they can only
occur in the same position as ``and'', which we treat as
a possible lexical realization of the anchor of the description-extending
auxiliary tree, but which is not presuppositional. It is not yet clear
to us which is the more appropriate way to treat them.

\subsection{Compositional vs. Inferential Semantics}
\label{compinf:sec}

One consequence of this approach is that clauses
linked by an explicit subordinate conjunction have a different
structural analysis than do clauses that are simply adjacent. This
might appear problematic because the perceived meaning of such
discourses is usually the same. For example,
\begin{enumerate}
\setcounter{enumi}{11}
\addtolength{\itemsep}{-5pt}
\item The City Council refused the women a permit because they
feared violence.
\item The City Council refused the women a permit. They
feared violence.
\end{enumerate}
\begin{figure*}
\centerline{\psfig{figure=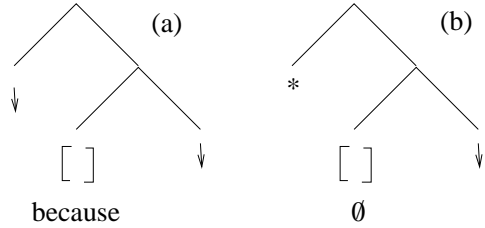}}
\caption{Trees used in the derivation of Ex.~12 and Ex.~13}
\label{because_tree:fig}
\end{figure*}
In our approach, (12) derives from the initial tree given in
Figure~\ref{because_tree:fig}a, while (13) derives from adjoining
an auxiliary tree (Figure~\ref{because_tree:fig}b)
to the tree for the first clause
and substituting the tree for the second clause at $\downarrow$.
Herein lies the difference between the two:
In (12), the causal connection is derived compositionally, while
in (13), one infers from the second utterance continuing the
description started in the first, that
the speaker intends the situation described in the second utterance to
serve as an explanation for that described in the first. Thus, the
causal connection is defeasible in (13) but not in (12).  This
can be seen by trying to continue each with ``But that wasn't the reason
for their refusal.'' The extended version of (12) seems ill-formed, while the
extended version of (13) seems perfectly coherent.

Another reason for distinguishing a limited compositional
semantics from an open inferential semantics is illustrated
by the following example:
\begin{enumerate}
\setcounter{enumi}{13}
\addtolength{\itemsep}{-5pt}
\item My car won't start. It may be out of gas.
\end{enumerate}
An RST analysis would simply decide what relation held between the
two clauses -- perhaps {\em non-volitional cause}. However,
non-volitional cause does not capture the different modal status
of the two clauses, which in turn affects the modal status of
the perceived relation: it is the car's {\em possibly}
being out of gas that is {\em possibly} the cause of its not
starting. We believe it is more systematic to just decide what
description is being continued (here, the one begun
in the first clause) and then derive further inferences that reflect
the different modal status of the two clauses. That the above inference
is defeasible can be seen by continuing the discourse in (14) with
``But that's not a possible reason for its not starting''.

\subsection{Brief Example}
\label{ex:sec}

Here we illustrate our approach by considering Example~9 (repeated
below) in more detail.
\begin{enumerate}
\setcounter{enumi}{8}
\addtolength{\itemsep}{-5pt}
\item John went to the zoo.
However, he took his cell phone with him.
\end{enumerate}
\begin{figure*}
\centerline{\psfig{figure=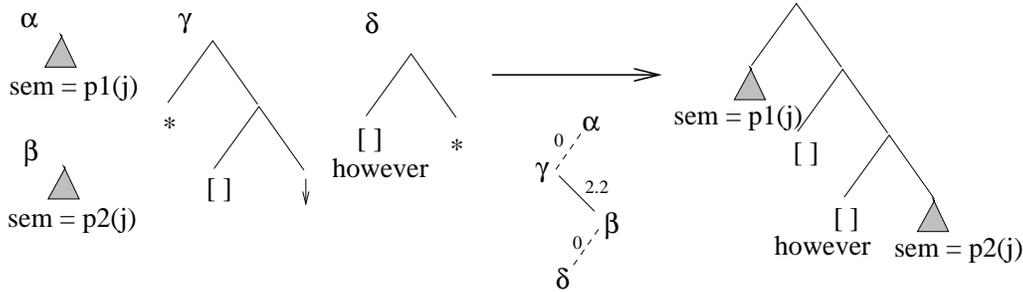}}
\caption{Derivation of Example 9}
\label{bugger:fig}
\end{figure*}
Three types of elements participate in the analysis: (1) the syntactic
analyses (trees) of the two clauses (``John went to the zoo, ``he
took his cell phone with him'') labelled $\alpha$ and $\beta$
in in Figure~\ref{bugger:fig}.,
along with their respective meanings (call them p$_1$(j) and  p$_2$(j));
(2) the auxiliary tree for the discourse cue ``however'', labelled
$\gamma$, along with its feature structure; and (3) the description-extending
auxiliary tree labelled $\delta$.

As the derivation in Figure~4 (below the arrow) shows, $\gamma$
adjoins at the root of $\alpha$, $\beta$ substitutes into
$\delta$, and $\delta$ adjoins at the root of $\beta$.
The semantics is as described earlier in Section~\ref{aux:sec}.

\section{Cue Phrases as Feature Structures (Knott, 1996)}
\label{feat:sec}

Earlier we noted that there was benefit to be gained from taking
the anchors of elementary trees to be feature structures into
which discourse cues (whose semantics was also in terms
of feature structures) could substitute. Here we briefly
argue why we believe this is so.

First, in viewing discourse cues in terms of feature structures,
we are following recent work by Knott
\shortcite{knott96,knott-mellish96}. Knott's study of the
substitutability patterns of discourse cues reveals that
their four common patterns -- synonymy, exclusivity,
hypernymy/hyponymy and contingent substitutability -- can, by
assuming inheritance (that, except for contingent substitutability,
a substitution pattern that holds for a discourse cue also holds for all
its hyponyms), follow from interpreting cues in feature-theoretic terms:
\begin{itemize}
\addtolength{\itemsep}{-5pt}
\item If cue $\alpha$ is {\em synonymous} with cue $\beta$, they signal the
same values of the same features.

\item If $\alpha$ is {\em exclusive} with  $\beta$, they signal
different values of at least one feature.

\item If $\alpha$ is a  {\em hypernym} of $\beta$, $\beta$ signals all
the features that $\alpha$ signals, as well as some features for
which $\alpha$ is undefined.

\item If $\alpha$ and $\beta$ are {\em contingently substitutable},
$\alpha$ and $\beta$ signal some of the same features, but
$\alpha$ is also defined for a feature for which $\beta$ is
undefined and $\beta$ is defined for a feature for which $\alpha$
is undefined.
\end{itemize}
Drawing on the extensive literature devoted to individual cue phrases,
Knott provided semantics for some of these features in terms of
{\em preconditions} on their use and/or
their {\em communicative effects}.

Following Knott in treating discourse cues in terms of feature
structures, it also appears beneficial to treat tree anchors
as feature structures as well, distinct from those of discourse
cues.

The reason for treating the anchor of subordinate clause initial
trees as feature structures is one of representational efficiency:
we can posit fewer such trees if we take their anchors to be
features structures that allow the (possibly contingent) substitution
of any subordinate conjunction with a compatible feature structure.
For example, we can have one tree whose anchor has the feature
{\em restricted-situation}, that can be realized as either ``if''
or ``when'' in some texts, but only ``when'' in others -- e.g.:
\begin{enumerate}
\addtolength{\itemsep}{-5pt}
\setcounter{enumi}{14}
\item Emergency parking regulations are in force $[$when, if$]$ more than
six inches of snow has fallen.
\item I found 30 new messages had arrived $[$when, *if$]$ I logged on
this morning.
\end{enumerate}
\cite{knott-mellish96} distinguish ``if'' and ``when'' by their
different values for the feature {\em modal status}: ``when'' has
the value {\em actual}, while ``if'' has the value {\em hypothetical}.
One can therefore say that other semantic features in Ex.~16 conflict
with the value {\em hypothetical}, only allowing ``when''. (N.B.
One could also take ``when'' as being unmarked for modal status,
its hypothetical reason begin synonymous with ``whichever''. The
conflict with ``if'' in Ex.~16 would still follow.)

The argument for treating the pair of anchors of parallel structures
as feature structures follows from the variability in the realization of
the medial anchor noted in Section~\ref{elem:sec}. One way to account
for this is that the anchor has features separate from those of
the discourse cues. Any cue
can then be used to realize the anchor, as long as it is either
\begin{itemize}
\addtolength{\itemsep}{-5pt}
\item less specific than the anchor, as in Ex.~3 -- ``but''
has few features in Knott's taxonomy;
\item more specific than the anchor, as in Ex.~5 -- ``on the other
extreme'', although it does not appear in Knott's taxonomy,  intuitively
appears to mean more than just ``side''.
\item partially overlapping with the anchor, as in Ex.~4 -- ``at the
same time'' 
has temporal features, but does not seem intrinsically contrastive. This
corresponds to Knott's concept of {\em contingent substitutability}.
\end{itemize}
It also appears as if the clause/discourse within the scope of an
anchor can either reinforce its features (as in Ex.~17 below) or
convey features of the anchor when it is not itself realized lexically,
as in Ex.~18:
\begin{enumerate}
\addtolength{\itemsep}{-5pt}
\setcounter{enumi}{16}
\item On the one hand, according to Fred, John is very generous.
On the other hand, according to everyone else, he will only give if
he sees an angle.
\item According to Fred, John is very generous.
According to everyone else, he will only give if
he sees an angle.
\end{enumerate}
But this part of our work is more speculative and the subject of
needed future work.

\section{Summary}

One way of seeing a grammar for discourse is as a story grammar -- i.e.,
a semantic grammar with components marked for the role they play in
the story or some sub-part. Alternatively, a discourse grammar
can, like a sentence-level
grammar, merely specify how structural units fit together and how the
semantics of the whole would be derived.
This is one such grammar.
While previous authors have adopted only certain aspects of TAG
or LTAG, here we have explored the possibility of a ``fully''
lexicalized TAG for discourse, which allows to examine how the
basic insights of a lexicalized grammar carry over to discourse.

Our proposal allows us to construct a smooth bridge between
syntactic clauses and discourse clauses,
each anchored on a lexical item (at times empty but always carrying the
appropriate features). It also allows us to factor out three separate
sources for elements of discourse meaning, thus providing a tool for
sorting out different processes in discourse and modeling them
individually. As such, we believe the approach provides some new insights
and tools for investigating discourse structure and discourse relations.

\smallskip
\begin{center}
{\large {\bf Acknowledgements}}
\end{center}

Our thanks to Mary Dalrymple, Christy Doran, Claire Gardent, Laura
Kallmeyer, Alistair Knott, Matthew Stone
and Mark Steedman for their invaluable comments and suggestions.
An earlier draft of this paper was presented at
the {\em Workshop on Underspecification}, Bad Teinach, Germany May 1998.

\end{document}